# Adverse Childhood Experiences Identification from Clinical Notes with Ontologies and NLP

Jinge Wu[1,2], Rowena Smith[2], Honghan Wu[1]
[1]Institute of Health Informatics, University College London, London, UK
[2]Usher Institute, University of Edinburgh, Edinburgh, UK

**Abstract**

*Adverse Childhood Experiences (ACEs) are defined as a collection of highly stressful, and potentially traumatic, events or circumstances that occur throughout childhood and/or adolescence. They have been shown to be associated with increased risks of mental health diseases or other abnormal behaviours in later lives. However, the identification of ACEs from free-text Electronic Health Records (EHRs) with Natural Language Processing (NLP) is challenging because (a) there is no NLP ready ACE ontologies; (b) there are limited cases available for machine learning, necessitating the data annotation from clinical experts. We are currently developing a tool that would use NLP techniques to assist us in surfacing ACEs from clinical notes. This will enable us further research in identifying evidence of the relationship between ACEs and the subsequent developments of mental illness (e.g., addictions) in large-scale and longitudinal free-text EHRs, which has previously not been possible.*

**Introduction**

Adverse childhood experiences (ACEs) encompass a broad variety of early traumas that occur in childhood, including both direct (e.g., abuse and neglect) and indirect (e.g., family mental illness and domestic violence) types [1, 2]. Numerous studies have established a substantial link between ACEs and negative health-related social outcomes. Liu et at. conducted systematic research including 2,129 studies and found that ACEs were consistently and positively associated with high rates of suicide (suicide attempts, major depression, substance abuse, adult victimization, etc.) [3]. Moreover, the lifetime frequency of ACEs is significantly greater in homeless individuals than in the general population, indicating ACE exposure may be related with an increased risk of mental illness, drug addiction, and victimization [3]. Policy actions and evidence-based interventions are urgently required to prevent ACE and to address the unfavourable consequences linked with this group of people.

The ACE causal route is complicated and most of such information is not coded in structured format, requiring the ability to interrogate multidimensional data including structured and unstructured Electronic Health Records (EHR) data. A standard vocabulary that facilitates data collection, analysis, interpretation, and sharing among varied parties and disciplines is crucial for enhancing the surveillance of ACEs. In this case, ontologies are regarded as standardized computational artifact used to represent knowledge in databases by defining concepts, instances, links, and axioms. They help increase interoperability of numerous data sources and systems, as well as data and information transmission across disciplines. They also enable the use of semantic technologies to detect new connections between datasets and hence new knowledge via logical inference. These ontologies have been widely used in health and biomedicine, and have significantly aided translational, clinical, and public health care delivery and management [5]. In order to promote the surveillance and study of adverse childhood experiences (ACEs), Brenas et al. standardized the ACE ontology and has been made publicly available as a formal reusable resource that can be utilised by the mental health research [4, 5]. The state-of-the-art ACEs Ontology (ACESO) has been available to the mental health community as well as the general public through the BioPortal data repository (https://bioportal.bioontology.org/ontologies/ACESO). Further in 2021, Ammar et al. [7] unveiled Semantic Platform for Adverse Childhood Experiences Surveillance (SPACES), a novel and explainable multimodal AI platform to assist ACEs surveillance, diagnosis of associated health issues, and subsequent therapies. Through the product, they generate suggestions and insights for better resource allocation and care management.

However, existing ontologies are not directly Natural Language Processing (NLP) applicable. There are at least two issues. First, ACE ontologies cover both high-level concepts (e.g., Disease or Findings) and concrete ACE concepts like child abuse. They do not provide information on which concepts are ACE specific, making it impossible to use the ontologies directly for NLP. Second, ACE ontologies do not provide synonyms and not all concepts are mapped to NLP-friendly terminologies like SNOMED-CT. Inspired by this, this study aims to make improvement on the previous ACE ontology by refining ACE concepts and its corresponding mappings to Unified Medical Language System (UMLS). Then, we use the refined terminologies with an off-the-shelf NLP tool to conduct a preliminary study on several open accessible corpus for identifying ACEs. Specifically, we choose SemEHR, an open-source toolkit for identifying mentions of UMLS concepts from textual data. This preliminary study will provide valuable experiences and identifying future research directions for our follow-up studies, aiming for developing efficient tools for extracting ACE evidence from clinical notes.

**Methods**

*1. Data collection from MIMIC and Reddit*

This project considers two datasets: MIMIC and Reddit, which can provide informative results from clinical and public health domain from EHRs and social media textual data.

Given that ACE patients are at high risk for serious illnesses like suicide, which will lead to ICU admissions, this project considers MIMIC-III dataset as part of the experiments. MIMIC-III [9] is one of the largest publicly available database containing sufficient free-text EHRs of ICU patients. It would be great to extract ACE related concepts for enhancement of the previous ontology. As a result, we used the discharge summaries in MIMIC-III, including 59,652 patients admitted to the ICU at Beth Israel Deaconess Medical Centre between 2001 and 2012.

Apart from clinical notes, the project also considers using social media textual data as complementation. The original Reddit dataset on mental health corpus [8] contains posts and text features from 826,961 unique users between 2018 and 2020 over 28 subreddits (15 mental health support groups). Due to time constraints, this paper only applied a subset of Reddit dataset with 2200 samples including 22 subreddits. The full dataset will be taken into account for future work.

*2. ACE ontology mapping*

The public ACE ontology, ACESO covers a wide range of ACEs containing 297 classes, 93 object properties, and 3 data properties with external links to SNOMED-CT and other ontology sources. However, there remain problems with ontology mappings, which lack consistent external links to standard clinical terminology systems. For example, we discovered there are only 112 ACE concepts linked with SNOMED-CT, leaving the rest unlinked. Besides, some of the linked concepts are too generic (eg. procedure) to extract ACE mentions from textual data using NLP tools.

To solve this problem, we contribute an approach by linking the ACE concepts to the widely used Unified Medical Language System (UMLS). Finding the leaf nodes from the ontology is an efficient way to obtain narrower ACE concepts. Therefore, we extract the leaf nodes from the ACESO and further construct one-to-one mapping to UMLS concepts. During the mapping, there might have multiple UMLS choice for each ACESO term, we can find the exact mapping in most cases. When it comes to ambiguous ones, we would manually choose the most relevant one.

*3. Identifying ACEs on MIMIC and Reddit using NLP tools*

We utilised SemEHR [6] for data extraction since it leverages UMLS concepts to find as many ACE mentions as feasible and contributes to disambiguation. To see how this works on different data sources, we use this NLP toolkit to identify ACEs from MIMIC and Reddit and listed the frequency of various ACE mentions. We also have a combination of results using project defined ACE terms only and combined terms (project defined terms and ACESO mapped terms).

**Experiments and results**

*1. Terminology for ACEs by combining UMLS and ACE ontology*

As a result, we extracted 140 unique leaf nodes out of 297 classes from ACESO concepts. Then the leaf nodes are mapped with UMLS concept CUI (eg. C0206073) using UMLS API match search. In total, 76 leaf nodes are able to find their corresponding UMLS concepts. Following that, a manual review was conducted aiming to remove irrelevant and high-level concepts, leaving 38 concepts in the final. In addition to the ACESO-defined terminologies, we also define extra 20 ACE concepts with links to UMLS for the project according to suggestions from domain experts. Our project-defined terms focus on narrow definition of ACEs and act as a good complement to ACESO concepts.

*2. Mentions in MIMIC and Reddit*

In each dataset, we conducted experiments using the project-defined ACE terminology and a combination of all ACE terminology (project-defined terms and ACESO mapped terms). Figure 1 shows ACE mentions from MIMIC data using all discharge summaries (n = 59,652). Figure 2 illustrates the ACE mentions from Reddit data including 2200 samples from 28 related subreddits (addiction, anxiety, depression, lonely, suicide_watch, etc.).

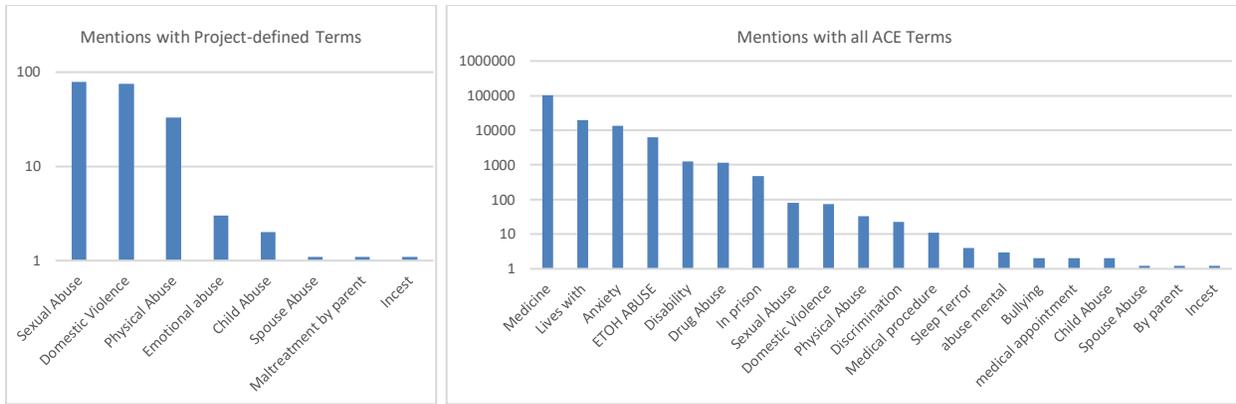

*Figure 1. ACEs identified in MIMIC data. In total, there are 195 mentions using project-defined terminology (20 concepts), and 145105 mentions using all ACE terminology (58 concepts).*

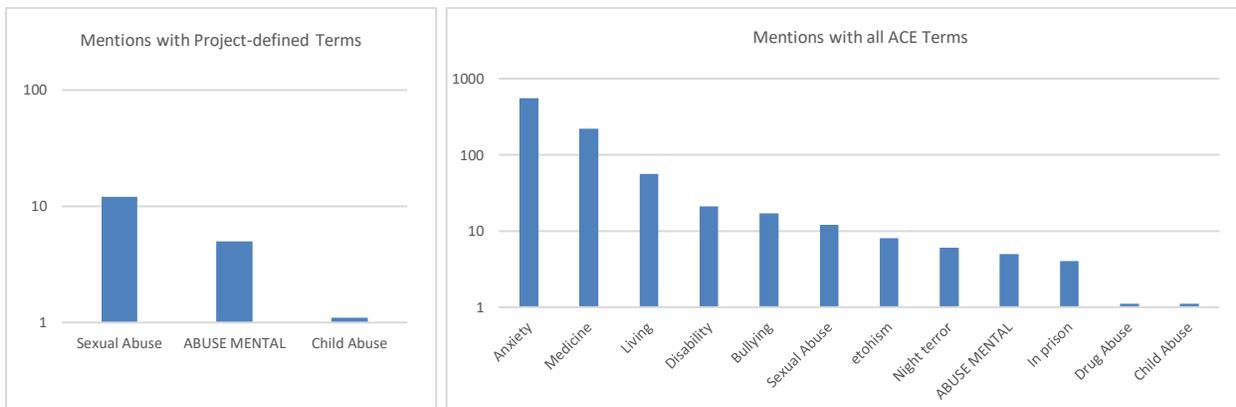

*Figure 2. ACEs identified in Reddit data. In total, there are 18 mentions using project-defined terminology (20 concepts), and 907 mentions using all ACE terminology (58 concepts).*

**Discussion**

According to the figures above, we observe that the current terminologies in general bring sufficient ACE mentions in the two datasets. Besides, our defined ACE terms act as a good complement as they identify a significant number of mentions from both datasets and are closely related to ACEs. This indicates our proposed method for combining UMLS and ACE ontology is efficient for NLP development.

Moreover, based on the results, we noticed that using the same terminologies brings distinct results for the two datasets. For MIMIC data, there is high proportional mentions of medical terms such as "Medicine" (102,563 mentions); while in terms of Reddit, the most frequent mention goes to "Anxiety" (557 mentions). This indicates that different terminologies might be needed for different types of datasets as "Medicine" is always one of the top mentions in EHRs but may not be efficient to identify ACEs.

However, it needs to be noted that the current ACE ontology still needs improvements. The reason for this is that, based on the combined terminology results, we observed a great number of mentions of "Medicine", "Anxiety" and "Living". These mentions might need further review as they can hardly be identified as ACEs. The second reason could be that there exists unmapped ACESO terms so that we could not use for NLP. Additionally, there are some issues with the concept selection such as "Household composition", leading to an increase in ambiguity and false positives. Therefore, in the future, more work should be done on defining clearer ACE concepts for different data sources and solving its ambiguous mappings to UMLS.

**Conclusion**

In summary, this paper provides valuable findings on improving existing ACE ontologies for better supporting clinical NLP. We proposed a practical and efficient approach to combine state-of-the-art ACE ontology (ACESO) with our project defined ACE terms with UMLS mappings. In addition, this paper identified ACEs from MIMIC and Reddit, leading to remarkable insights from both clinical and social media domain. As there still requires improvements, future steps also involve 1) investigate a scalable and efficient way to improve ACE terminologies by integrating existing ontologies; 2) refinement of the ACEs concepts to solve the mapping problem with supports from domain experts; and 3) create a public accessible benchmark on MIMIC/Reddit datasets for ACE identification with gold-standard annotations by domain experts.

**References**

1. Hughes K, Bellis MA, Hardcastle KA, Sethi D, Butchart A, Mikton C, Jones L, Dunne MP. The effect of multiple adverse childhood experiences on health: a systematic review and meta-analysis. The Lancet Public Health. 2017 Aug 1;2(8):e356-66.
2. Felitti VJ, Anda RF, Nordenberg D, Williamson DF, Spitz AM, Edwards V, Marks JS. Relationship of childhood abuse and household dysfunction to many of the leading causes of death in adults: The Adverse Childhood Experiences (ACE) Study. American journal of preventive medicine. 1998 May 1;14(4):245-58.
3. Liu M, Luong L, Lachaud J, Edalati H, Reeves A, Hwang S. Adverse childhood experiences and related outcomes among adults experiencing homelessness: a systematic review and meta-analysis. The Lancet Public Health. 2021;6(11):e836-e847.
4. Brenas JH, Shin EK, Shaban-Nejad A. An Ontological Framework to Improve Surveillance of Adverse Childhood Experiences (ACEs). InEFMI-STC 2019 Jan 1 (pp. 31-35).
5. Brenas JH, Shin EK, Shaban-Nejad A. Adverse Childhood Experiences Ontology for Mental Health Surveillance, Research, and Evaluation: Advanced Knowledge Representation and Semantic Web Techniques. JMIR Mental Health. 2019;6(5):e13498.
6. Wu H, Toti G, Morley KI, Ibrahim ZM, Folarin A, Jackson R, Kartoglu I, Agrawal A, Stringer C, Gale D, Gorrell G. SemEHR: A general-purpose semantic search system to surface semantic data from clinical notes for tailored care, trial recruitment, and clinical research. Journal of the American Medical Informatics Association. 2018 May;25(5):530-7.
7. Ammar N, Zareie P, Hare M, Rogers L, Madubuonwu S, Yaun J et al. SPACES: Explainable Multimodal AI for Active Surveillance, Diagnosis, and Management of Adverse Childhood Experiences (ACEs). 2021 IEEE International Conference on Big Data (Big Data).
8. Low DM, Rumker L, Talkar T, Torous J, Cecchi G, Ghosh SS. Natural language processing reveals vulnerable mental health support groups and heightened health anxiety on reddit during covid-19: Observational study. Journal of medical Internet research. 2020 Oct 12;22(10):e22635
9. Johnson AE, Pollard TJ, Shen L, Lehman LW, Feng M, Ghassemi M, Moody B, Szolovits P, Anthony Celi L, Mark RG. MIMIC-III, a freely accessible critical care database. Scientific data. 2016 May 24;3(1):1-9.